\documentclass{article}

\pdfoutput=1
     \PassOptionsToPackage{numbers, compress}{natbib}

     \usepackage[preprint]{neurips_2019}

\usepackage[utf8]{inputenc} 
\usepackage[T1]{fontenc}    
\usepackage{url}            
\usepackage{booktabs}       
\usepackage{amsfonts}       
\usepackage{nicefrac}       
\usepackage{microtype}      

\usepackage{graphicx}
\usepackage{wrapfig}
\graphicspath{{figures/}}

\usepackage{subfigure}
\usepackage{booktabs} 

\usepackage{algorithm}
\usepackage{algorithmic}

\usepackage{verbatim}
\usepackage[font=footnotesize]{caption}
\usepackage{enumitem}
\usepackage{color}
\usepackage{multirow}

\newcommand{\redVspace}{\vspace{-4mm}}

\title{Coloring Big Graphs With AlphaGoZero}

\author{%
  Jiayi Huang\stepcounter{footnote}\thanks{Work done at SVAIL, Baidu, USA.
  Correspondence to: Gregory Diamos <gregory.diamos@gmail.com>.}\\
  Texas A\&M University\\
  \texttt{jyhuang@tamu.edu}
  \And
  Mostofa Patwary\footnotemark[2]\\
  NVIDIA\\
  \texttt{mostofa.patwary@gmail.com}
  \And
  Gregory Diamos\footnotemark[2]\\
  Landing AI\\
  \texttt{gregory.diamos@gmail.com}
}

\begin{document}

\maketitle

\begin{abstract}
  Recent innovations in deep reinforcement learning have shown promising potential to develop improved heuristics for complex problems.
  Graph coloring, as a well-known NP-hard problem with clear commercial applications, requires specialized knowledge to design well-performed heuristics for different problems.
  In this paper, we adapt AlphaGoZero with graph embedding to learn better heuristics for graph coloring with zero prior knowledge using high performance computing technologies.
  Key to our approach is the introduction of a novel deep neural network architecture (FastColorNet) that has access to the full graph context and requires $O(V)$ time and space to color a graph with $V$ vertices, which enables scaling to very large graphs that arise in many real applications. We show that our approach can effectively learn new state-of-the-art heuristics for graph coloring.

\end{abstract}

\section{Introduction}


Current approaches for quickly solving NP-hard optimization problems like graph coloring rely on carefully designed heuristics.  These heuristics achieve good general purpose performance, and are fast enough to scale up to very large problems.  However, the best performing heuristic often depend on the problem being solved. Recently, machine learning methods have shown promising potential to develop improved heuristics for specific application domains~\cite{alpha-go-zero, structure2vec-q-learning, gcn-guided-tree-search, learned-auto-reason-heuristic}.   

In order to learn strong heuristics for specialized applications of graph coloring, we need training data that includes solutions that outperform existing heuristics.  However, the extremely large search spaces of even single instances of NP-hard problems like graph coloring presents a significant challenge. Recently, deep reinforcement learning algorithms have found recent success in games like Go, Chess, and Shogi~\cite{alpha-zero} with very large search spaces.  These algorithms use deep neural networks to store knowledge learned during self-play. They build on this knowledge using search procedures such as Monte Carlo Tree Search (MCTS) with Upper Confidence Bound (UCB), leading to better solutions with more training. Although they require many training steps to achieve good performance, the inexpensive evaluation of solutions to NP-hard problems enables very fast training.  



%

\begin{figure}[thb]
\begin{center}
\includegraphics[width=\textwidth]{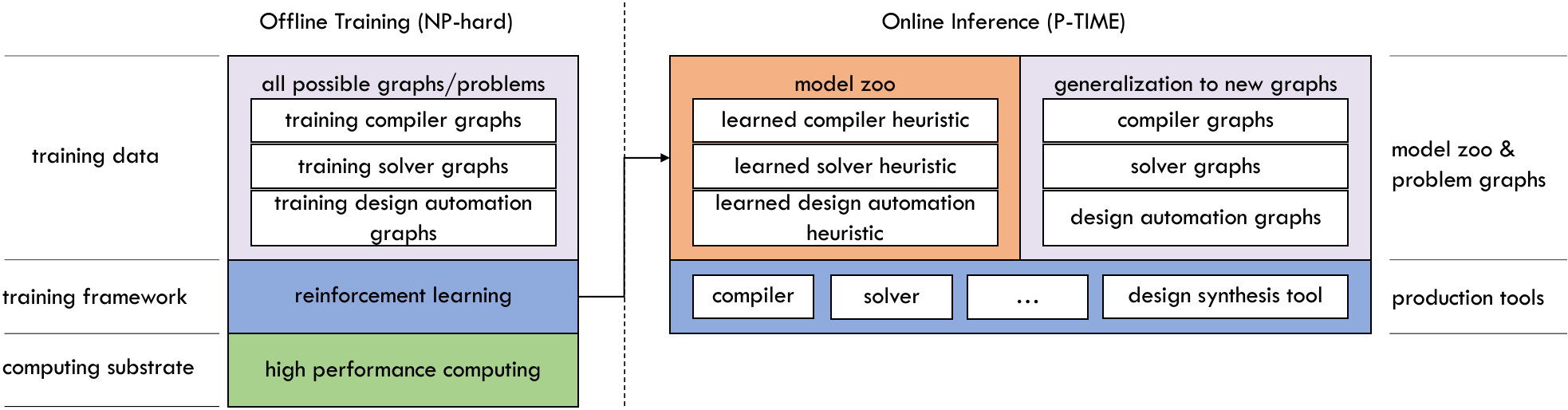}
  \caption{\textbf{Deep Heuristic Learning}.  High performance training systems use deep reinforcement learning to improve heuristics offline, which are then deployed online in production tools.} 
\label{fig:heuristic-learning-framework}
\end{center}
\end{figure}

In this paper we demonstrate our approach by introducing a framework for learning new heuristics using deep reinforcement learning, depicted in Figure~\ref{fig:heuristic-learning-framework}.
A reinforcement learning algorithm is used to extend the performance of the best existing heuristic on a training set of problems.  
Training is still NP-hard, even on a single problem.  However, high performance computing (HPC) systems can devote large scale computation and significant training time to building stronger heuristics.
The best heuristics discovered during training may be distilled into models that are fast to evaluate (e.g. P-TIME, parallel, etc), and stored in a model zoo.  These heuristics could be periodically downloaded into production tools, which would then be capable of quickly finding good solutions to the same problems that were encountered during training.  We seek to address the following question: ``how well do learned heuristics generalize, and for which application domains"?

Our results suggest that a similar approach can be successfully applied to other combinatorial optimization problems.
Additionally, our results can be further improved by using even faster training systems to run deep reinforcement learning on larger datasets of representative problems.
The specific contributions of this paper are as follows:
\begin{itemize}[noitemsep, topsep=0pt, leftmargin=2em]
    \item We introduce a framework for learning fast heuristics using deep reinforcement learning algorithms inspired by AlphaGoZero~\cite{alpha-go-zero} on HPC systems, which is used to learn new graph coloring heuristics that improve the state-of-the-art accuracy by up to $10\%$
    \item We demonstrate how to express the graph coloring problem as a Markov Decision Process in and apply the self-play reinforcement learning algorithm in AlphaGoZero to graph coloring.
    \item We introduce a new highly optimized neural network FastColorNet in Section~\ref{sec:fast-color-net} that has access to the full graph context, exposes significant parallelism enabling high performance GPU implementations, and can scale up to very large graphs used in production tools, which have many order of magnitude (${\sim}100K$) larger search space than Go.
\end{itemize}

\section{The Graph Coloring Task}

\begin{wrapfigure}{r}{0.45\columnwidth}
  \vspace{-2em}
  \begin{tabular}{cc}
    \includegraphics[width=0.2\columnwidth]{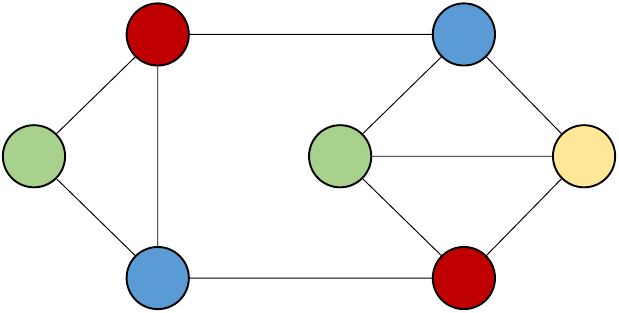} &
    \includegraphics[width=0.2\columnwidth]{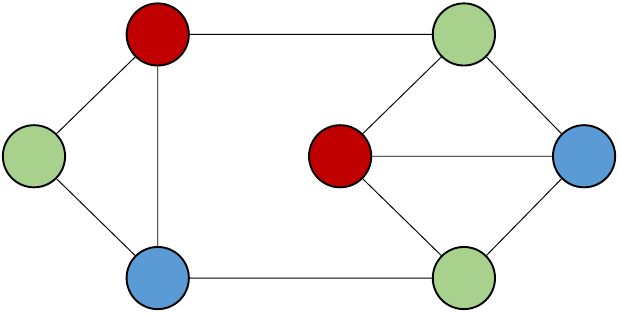} \\
    (a) \footnotesize{4 colors} &
    (b) \footnotesize{3 colors} \\
  \end{tabular}
  \caption{A \textit{slightly-hard-to-color} graph~\cite{smallest-hard-to-color-graph} is colored with greedy heuristics (static-order, sorted-order, and dynamic-sorted-order heuristics) in (a) and colored optimally in (b).}
\label{fig:colored-bucky}
\end{wrapfigure}

Graph coloring as illustrated in Figure~\ref{fig:colored-bucky} is one of the core kernels in combinatorial scientific
computing with many practical applications including parallel
computing, VLSI, and pattern matching \cite{naumann2012combinatorial}. It assigns colors to the vertices of a given graph $G=(V, E)$ such that no adjacent vertices are of the 
same color. The objective of the graph coloring problem is to 
minimize the number of colors used, and known to NP-hard to solve 
optimally \cite{gebremedhin2005color}. In fact, graph coloring is 
even NP-hard to approximate in specific scenario \cite{zuckerman2006linear}.
Therefore, linear time greedy algorithms are often used in practice, 
which yield near optimal solutions in practice \cite{ccatalyurek2012graph}. It is worthwhile to mention that the 
order of vertices used in the greedy coloring algorithm plays an
important role on the quality of the solutions \cite{gebremedhin2013colpack}.

In this paper, we use matrix $C$ to represent the assignment of colors to graph $G$ where $C_{i,j}$ 
indicates whether a unique color $j$ is assigned to vertex $i$ in $G$. The above mentioned greedy coloring heuristics
hence progressively update $C$, each vertex at a time until all vertices in $G$ are 
assigned colors.  




\subsection{Graph Coloring as a Markov Decision Process}

In order to apply deep reinforcement learning to the graph coloring problem, we represent graph coloring as a Markov Decision Process (MDP).  Here, $C^{(t)}$ encodes the MDP state $s^{(t)}$ at step $t$.  Recall that $C^{(t)}$ represents the assignment of colors to vertices at step $t$.  The set of actions, $A_i$, is the set of valid colors that may be assigned to the next vertex $i$ at step $t$.  All actions are deterministic, and the intermediate reward $R_a$ is the negative total number of colors used so far.
Our goal is to learn {\boldmath$\pi$}$_*(s)$, the optimal policy mapping from state $s^{(t)}$ to action $a_i$, and $V(s^{(t)})$ the expected reward of $s^{(t)}$ while following {\boldmath$\pi$}$_*$.
It is important to recognize that unlike games of the Go or Chess, graph coloring heuristics need to support diverse graphs, and different graphs imply different MDPs.  

\subsection{A Zero Sum Game of Graph Coloring}

One difference between graph coloring and zero sum games like Go and Chess is the reward.  In graph coloring the most obvious choice of reward is the number of colors used.  In games like Chess and Go, the reward is typically win, lose, or in some cases tie.  In these games it is natural for algorithms to exploit self-play, where the best performing learned algorithm plays against itself.

We experiment with a new reward for graph coloring inspired by self-play.  We use the best learned algorithm so far to color the graph.  We define a new reward for any solution with fewer colors to win, any solution with more colors to lose, and any solution with the same number of colors to tie.  We find that this choice simplifies the design of reward scaling and alpha-beta-pruning, although both reward formats are able to achieve comparable results.

\subsection{Is Graph Coloring Harder Than Go?}

To put graph coloring in context with other deep reinforcement learning tasks, we estimate the size of the MDPs in terms of number of states for graph coloring as well as Go and Chess by multiplying the average number of moves by the average number of actions per move on a set of graphs.  Table~\ref{table:search-state-sizes} shows that even moderately sized graphs imply very large MDPs, and the largest graphs imply MDPs that are several order of magnitude larger than Go.  As we will see, these much larger problems present significant scalability challenges in training and inference.
In addition, graph coloring has a unique challenge of state representation for supporting diverse graphs of different sizes and structures compared to Go that has static-dimension board view. 
In this work, we extend the AlphaGoZero approach with scalable graph embedding to address them.



\begin{table*}[th]
  \vspace{0.5em}
\begin{center}
  \caption{Estimated MDP states for \textbf{single} random graphs of various sizes compared to Chess and Go.}
  \vspace{-0.5em}
\label{table:search-state-sizes}
  \resizebox{\textwidth}{!}{
\begin{tabular}{|l|l|l|l|l|l|l|l|}
\hline
                 & Graph-32 & \textbf{Chess} & Graph-128 & \textbf{Go}    & Graph-512 & Graph-8192 & Graph-$10^7$ \\ \hline
Avg. MDP States & $10^{21}$    & $\mathbf{10^{60}}$  & $10^{141}$    & $\mathbf{10^{460}}$ & $10^{790}$    & $10^{19,686}$   & $10^{45,830,967}$     \\ \hline
Avg. Moves Per Game & 32    & \textbf{40}  & 128    & \textbf{200} & 512    & 8,192   & $10^7$     \\ \hline
\end{tabular}
}
\end{center}
  \vspace{-1em}
\end{table*}

\section{Graph Coloring with AlphaGoZero}
\label{sec:graph-coloring-with-alphagozero}

\begin{figure}[ht]
\begin{center}
\centerline{\includegraphics[width=0.8\columnwidth]{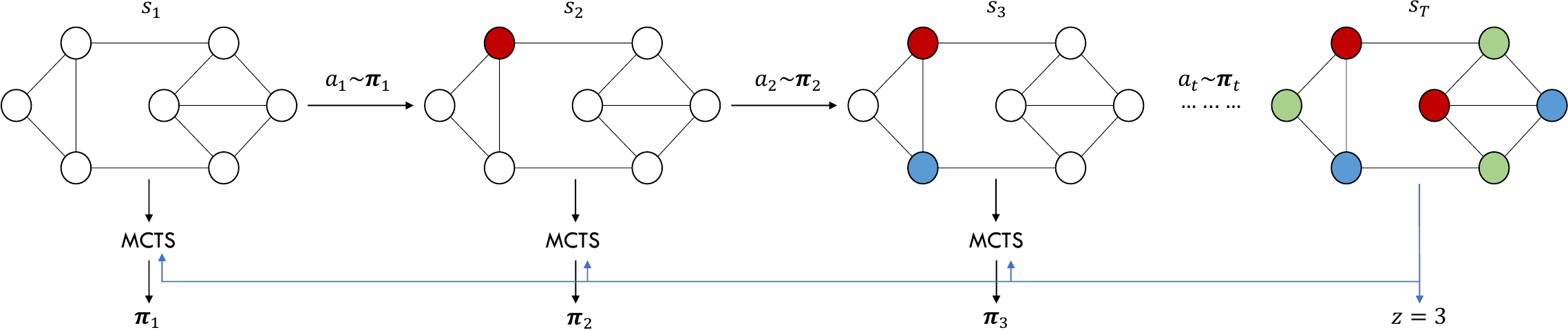}}
  \caption{The reinforcement learning algorithm.  At each state, a MCTS computes probabilities {\boldmath$\pi$}$_i$ for the current move, and the next action $a_i$ is selected. When the graph is colored, the final score $z$ is stored as a label for all previous moves. }
\label{fig:alpha-go-zero}
\end{center}
\vskip -0.2in
\end{figure}

In this section we show how AlphaGoZero can be adapted and applied to graph coloring.  
We use a deep neural network $f_\theta$ with parameters $\theta$ to map a representation $s$ of the graph state and coloring history to next-color probabilities and a value, $(\mbox{\boldmath$p$},v) = f_\theta(s)$.  
{\boldmath$p$} represents the probability of selecting each valid color next, and $v$ is a scalar estimating the probability of finding a better final solution than the current best heuristic from position $s$.  
We use MCTS+UCB to search for better labels ({\boldmath$\pi$}, $z$) as shown in Figure~\ref{fig:alpha-go-zero}, which the neural network can then be trained on, resulting in an even stronger neural network.  
This process is then repeated in a policy iteration procedure.  

\subsection{Monte Carlo Tree Search}

\begin{figure}[ht]
\begin{center}
\centerline{\includegraphics[width=0.8\columnwidth]{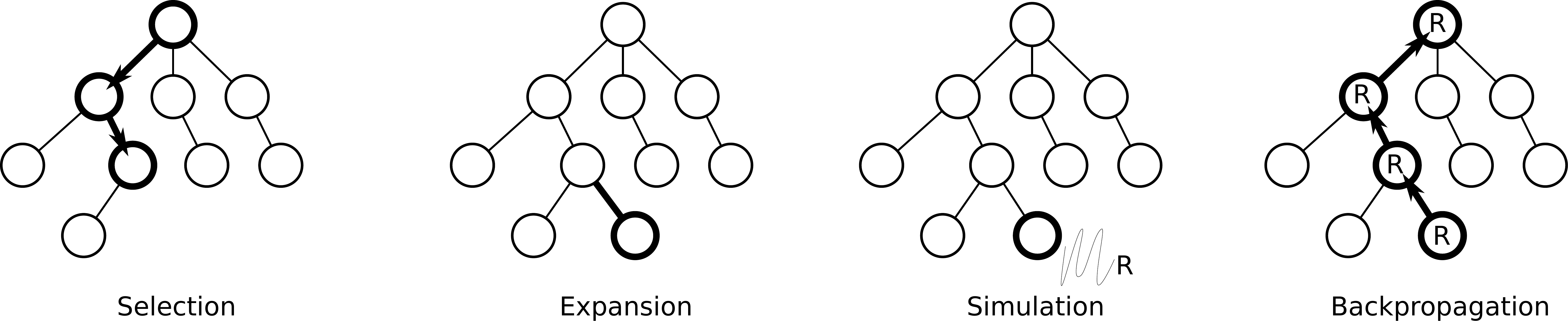}}
\caption{The four stages of the Monte-Carlo Tree Search algorithm.  A path is \textbf{selected} that maximizes the UCB, the tree is \textbf{expanded}, the neural network performs a \textbf{simulation}, and the result is \textbf{back-propagated} to update nodes along the current path.}
\label{fig:monte-carlo-tree-search}
\end{center}
\vskip -0.2in
\end{figure}

Following the implementation in~\cite{alpha-go-zero}, Monte-Carlo tree search uses the neural network $f_\theta$ as a guide.   The search tree stores a prior probability $P(s, a)$, a visit count $N(s, a)$, and a state action-value $Q(s, a)$ at each edge $(s, a)$.  Starting from the root state, the next move is selected that maximizes the upper confidence bound 
$Q(s, a) + U(s, a)$ until a leaf node $s_l$ is reached as shown in the \textbf{Selection} panel of Figure~\ref{fig:monte-carlo-tree-search}.  The leaf node $s_l$ is expanded as shown in the \textbf{Expansion} panel of Figure~\ref{fig:monte-carlo-tree-search}, and is evaluated by the neural network to compute $(P(s_l,*), V(s_l)) = f_\theta(s_l)$, corresponding to the \textbf{Simulation} panel of Figure~\ref{fig:monte-carlo-tree-search}. Each edge visited during this process is updated to increment its visit count $N(s, a)$ and set its action-value to the mean value $Q(s, a) = (Q(s, a) * N(s, a) + V(s_l))/(N(s, a) + 1)$.  This is shown in the \textbf{Backpropagation} panel of Figure~\ref{fig:monte-carlo-tree-search}.  

\subsection{Upper Confidence Bound}

A core issue in reinforcement learning search algorithms is maintaining the balance between the exploitation of moves with high average win rate and the exploration of moves with few simulations.   
We follow the variant of UCB used in~\cite{alpha-go-zero} for computing $U(s, a)$ and handling this tradeoff.

\vspace{-0.2in}
\begin{equation}
    U(s, a) = c * P(s, a) * \frac{\sqrt{\sum_{i=b}^{M}{N(s,b))}}}{1 + N(s, a)}
\end{equation}
\vspace{-0.2in}

Here $M$ is the number of available actions at the current state, and $c$ is the exploration factor hyperparameter, typically set between 0.1 and 3.0.  
The combined MCTS+UCB algorithm initially focuses on moves with high prior probability and low visit count (exploration), and eventually prefers moves with high $Q(s, a)$ value (exploitation).

\subsection{Self-Play}

To generate labels ({\boldmath$\pi$}, $z$) for each move in a new graph $G$, the graph is first colored by running the neural network of best learned heuristic so far over the entire graph, producing a baseline score $\chi(G)$.
MCTS+UCB is then used to color the same graph and produce a comparing score.
The existing search tree along the selected move is reused for the next move and new search probabilities {\boldmath$\pi$} are computed.
Sampling is used for the first several moves to encourage exploration (controlled by a hyperparameter), then max decoding is used for subsequent moves.  
We use alpha-beta-pruning~\cite{alpha-beta-pruning} to abort plays early that are clearly won or clearly lost compared to the baseline score $\chi(G)$.  
Results from self play are stored as tuples ($G$, $C$, {\boldmath$\pi$}, $z$), one for each move. The neural network is trained by sampling moves uniformly from a replay buffer~\cite{replay-buffer} of the most recent moves.  The replay buffer stores a single copy of the graph for all moves on that graph to save memory.  It also lazily materializes a cache of embeddings for the same reason.  

Games of Go typically end after a few hundred moves, but some of our graphs have millions of nodes.
Running MCTS all the way to the end for each move is not feasible.
Furthermore, given the computational cost of training deep neural networks, it is only feasible to train on about 1 billion moves in a reasonable amount of time, even on high performance clusters of accelerators.
For example, AlphaGoZero trained on a TPU cluster for about 1.4 billion moves. Given graphs with tens of millions of nodes, this would only allow training on about 100 games of self-play.
To scale for very large graphs, we add two additional techniques to the self-play algorithm.
\begin{itemize}[noitemsep, topsep=0pt, leftmargin=2em]
  \item \textbf{Limited-Run-Ahead.} We restrict the MCTS to a limited number of future moves that is typically set to several hundred.
    To evaluate $z$, we compare the score of the baseline after run-ahead-limit with the score produced by MCTS.
  \item \textbf{Move-Sampling.} To generate more diversity of labels, we only choose a subset of all moves sampled uniformly from all moves in all graphs in the training set to apply MCTS and to produce training samples.
    After choosing a move, we run MCTS for several consecutive moves to avoid resetting the search.
    We use a fast (smaller) model to fast-forward to the next sampled move.
\end{itemize}

\subsection{FastColorNet}
\label{sec:fast-color-net}

\begin{figure}[ht]
\begin{center}
\centerline{\includegraphics[width=0.8\columnwidth]{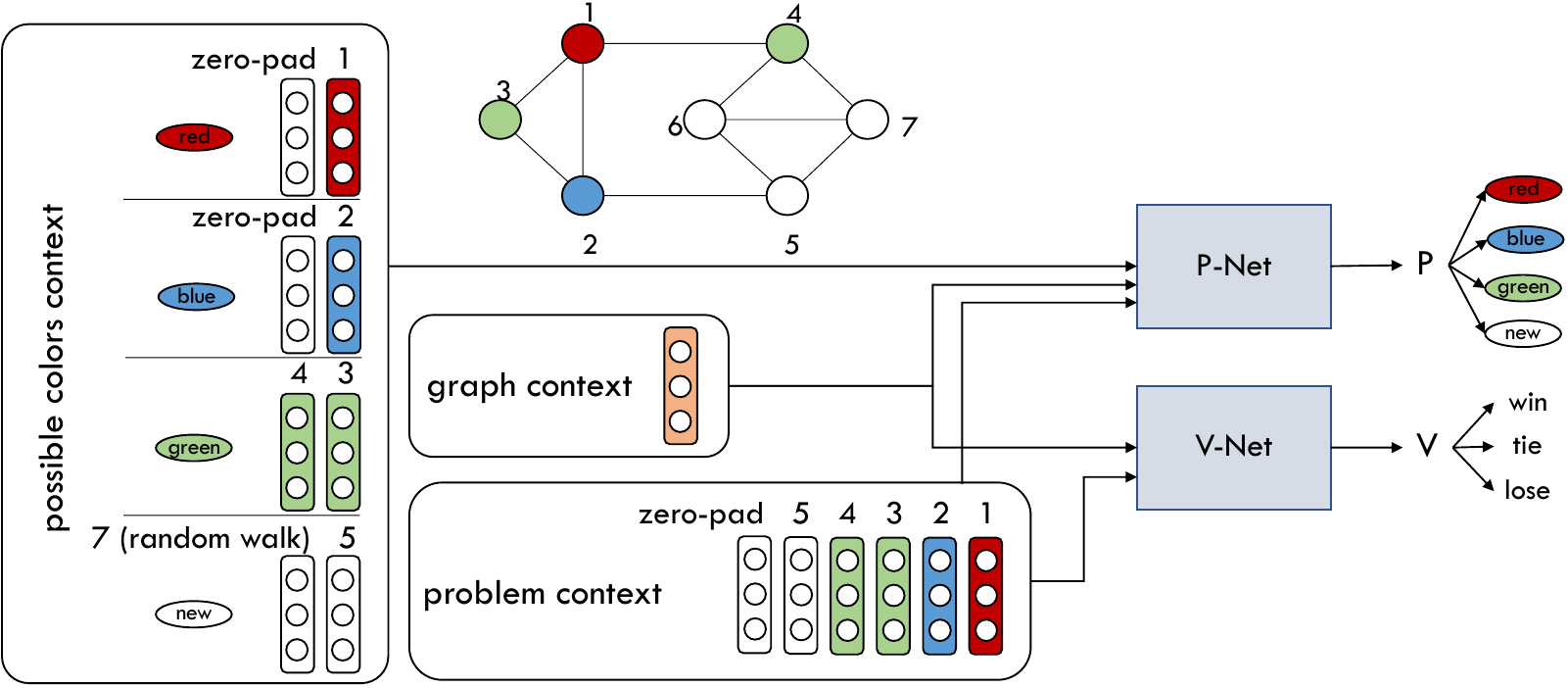}}
  \caption{Our FastColorNet architecture computes $(\mbox{\boldmath$p$}, v)$ for graph $G$ and colors $C$.  The numbered vertices represent the coloring order immediately before and after the current move.}
\label{fig:fast-color-net}
\end{center}
\redVspace
\redVspace
\end{figure}

Heuristics are used in place of exact solvers because they are fast. In order to compete, we need a neural network that can keep up.  This leaves us with the following goals:
\begin{itemize}[noitemsep, topsep=0pt, leftmargin=2em]
\item \textbf{Scalability}.
Fast graph coloring heuristics run in linear $O(V)$ or $O(E + V log V)$ time, enabling scaling to big graphs. For example, some graphs in SuiteSparse have over 10 million vertices.
\item \textbf{Full Graph Context}.
We expect different graphs to require different coloring strategies, so our network needs information about the structure of the graph being colored.
\end{itemize}

FastColorNet accomplishes both goals by i) using a scalable message passing algorithm for computing node embeddings that can capture the global graph structure, and ii) performing a dynamically sized softmax that assigns a probability of selecting each of the valid colors for the current node.  


The complete FastColorNet architecture is shown in Figure~\ref{fig:fast-color-net}.  The network takes graph embeddings and a few simple features such as the number of vertices in the graph as inputs. It predicts $V$, the expected result of coloring the remainder of the graph, and $P$, a distribution over available colors for the current node.  Note that $P$ is variably sized.  FastColorNet is trained end-to-end from $V$ and $P$ labels to embeddings.  It used stochastic selection of embeddings and an extension of truncated back propagation through time to keep the computational requirement to color each vertex constant for both training and inference.

We note that FastColorNet represents the first architecture that we were able to design that fulfills all of these requirements while being able to consistently fit the $V$ and $P$ labels produced by the MCTS.  We believe that there is room for future work to significantly improve the accuracy of this model with more extensive architecture search.

\subsubsection{Graph Embeddings}

\begin{wrapfigure}{R}{0.5\textwidth}
\vspace{-3em}
  \begin{minipage}{0.5\textwidth}
    \begin{algorithm}[H]
      \footnotesize
  \caption{Graph Embedding}
  \label{alg:graph-embedding}
  \begin{algorithmic}[1]
    \STATE {\bfseries Input:} parameters $\mathbf{\theta} \in \widetilde{\mathcal{T}}$
    \STATE Initialize $\widetilde{\mu}_{i}^{(0)} = \textbf{0}$, for all $i \in  \mathcal{V}$
    \FOR{$t=1$ {\bfseries to} $T$}
      \FOR{$i \in \mathcal{V}$}
        \STATE $\nu_{i} = [d_{i},\widetilde{\mu}_{i}]$, $d_{i}$ is $i$'s degree (one-hot)
        \STATE $l_{i} = [\nu_{i}, d_{j},\widetilde{\mu}_{j}^{(t-1)}]$, where $j$ = random($\mathcal{N}(i)$)
        \STATE $c_i = 0$ 
        \FOR{$k=1$ \textbf{to} $L$}
          \STATE{$l_{i},c_i = LSTM_{\theta}(c_i, l_i)$}
        \ENDFOR
        \STATE{$\widetilde{\mu}_{i}^{(t)} = LSTM_{\theta}(c_i, v_i)$} 
      \ENDFOR
    \ENDFOR{\{fixed point equation update\}}
    \STATE{return ${\{\widetilde{\mu}_{i}^{T}\}_{i \in \mathcal{V}}}$}
  \end{algorithmic}
\end{algorithm}
  \end{minipage}
\vspace{-3em}
\end{wrapfigure}

In order to compute $V$, information $G$ and $C$ is required.  Simply passing $G$ and $C$ would be prohibitively expensive and be unable to generalize to instances of $G$ and $C$ with different sizes. We provide information to FastColorNet through graph embeddings, which are defined per vertex, and are intended to learn relevant local and global information about $G$ and $C$.

Designing graph embeddings that capture local and global graph structure is challenging given the large ranges of sizes and structures that must be supported with a fixed size vector~\cite{embeddings-survey}.  
Our design is inspired by ~\cite{message-passing-embeddings, sampled-embedding-training}'s work extending message passing algorithms such as loopy belief propagation to graph embeddings.  

Like structure2vec~\cite{message-passing-embeddings}, we start from near-zero initialized embeddings and run loopy belief propagation using a learned transfer function $\widetilde{\mathcal{T}}$.  Intuitively, neighboring vertices exchange messages which are processed by transfer function $\widetilde{\mathcal{T}}$.  After enough iterations, this series of messages is likely to converge to a consensus.  
Like \cite{sampled-embedding-training}, we seek to avoid running back propagation through the entire graph each time an embedding is used, and instead take a sampling approach.  Using an analogy of truncated back propagation~\cite{truncated-back-prop} applied to graphs, we apply back propagation only along a random walk ending at each referenced vertex embedding.  

We represent the transfer function $\widetilde{\mathcal{T}}$ as a neural network with weights that are learned using back propagation through the embeddings.  We experimented with several architectures for the $\widetilde{\mathcal{T}}$ and found training to be more stable with an LSTM than the fully connected network in~\cite{message-passing-embeddings}.

During inference, we compute embeddings for each vertex by iteratively applying the transfer function to vertices.  This approach naturally allows batching up to the size of the graph, making it extremely compute intensive and suitable for GPU or distributed acceleration. The total amount of computation required can be controlled by the number of iterations.  In our experiments we run for three iterations. The algorithm is described in Algorithm~\ref{alg:graph-embedding}.

\subsubsection{Inputs and Outputs}

FastColorNet predicts $(\mbox{\boldmath$p$}, v)$ for the next move, and is trained on labels ({\boldmath$\pi$}, $z$) from the MCTS.  Recall that $V$ can be one of the following labels (win, tie, lose), and that {\boldmath$\pi$} has one entry for each valid color that has already been assigned in addition to a label corresponding to allocating a new color. 
The loss is the sum of individual cross entropy losses: 

\vspace{-0.2in}
\begin{equation}
  L(\mbox{\boldmath$\pi$}, \mbox{\boldmath$p$}, z, v) = \mbox{\boldmath$\pi$}^T log(\mbox{\boldmath$p$}) + z^T log(v)
\end{equation}
\vspace{-0.2in}


Graph context shown in the middle of Figure~\ref{fig:fast-color-net} contains the following one-hot encoded values (the number of vertices in G, the total number of assigned colors, the number of vertices that have already been colored) concatenated with multi-hot encoded set of valid colors for the current vertex.

Problem context shown in the bottom of Figure~\ref{fig:fast-color-net} contains the embeddings of vertices that have just been colored, and vertices are scheduled to be colored next, in order.  Problem context is zero padded at the beginning and end of the graph.  The number of vertices included in the problem context is a hyperparameter that we typically set to 8 in experiments.

Possible colors context shown in the left of Figure~\ref{fig:fast-color-net} contains the embeddings of fixed size sets of vertices that have been colored with each of the possible colors.  Possible colors context is zero padded if not enough vertices have been assigned to fill out a complete set for one of the possible colors.  The set size is a hyperparameter that we typically set to 4 in experiments.

\begin{figure}[htb]
  \centering
  \subfigure[V-Network]{
    \includegraphics[height=0.3\textwidth]{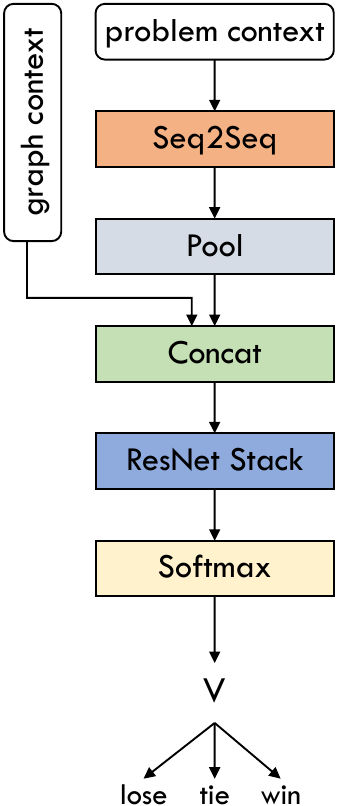}
    \label{fig:v-network}
  }
  \qquad
  \subfigure[P-Network]{
    \includegraphics[height=0.3\textwidth]{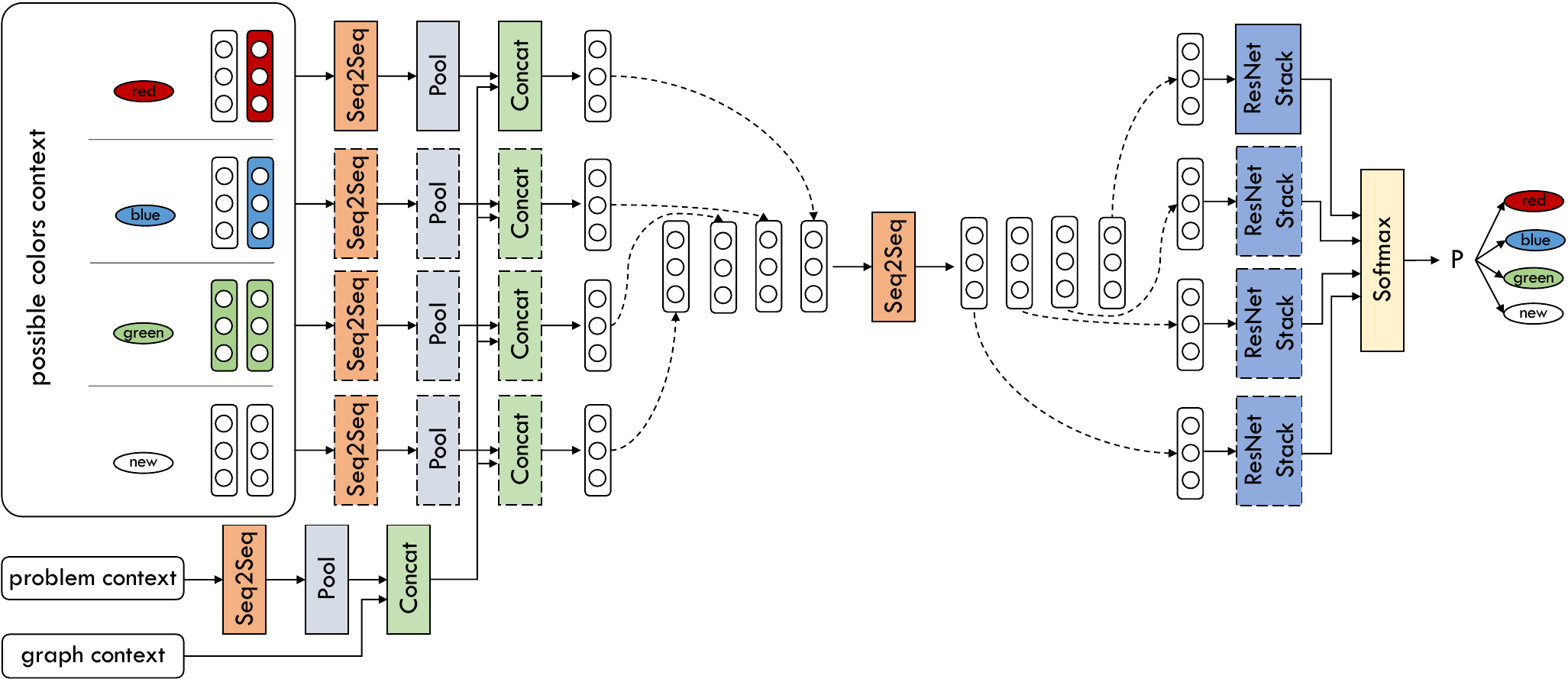}
    \label{fig:p-network}
  }
  \caption{The V-Network estimates the outcome of the coloring problem from the current state using graph context and problem context (a) and the P-Network evaluates the probability of possible valid colors for the current node using graph context, problem context and possible colors context.}
  \label{fig:network}
\end{figure}

\subsubsection{V-Network}

The V network architecture shown in Figure~\ref{fig:v-network} maps from the graph context and the problem context to the expected outcome of the coloring problem.  The vertex embeddings in the problem context are stored in a sequence corresponding to the order that vertices are colored.  We use a sequence-to-sequence model to exploit locality in this sequence, followed by a pooling operation over the sequence to summarize this information.  We typically use stacks of residual 1-D convolution layers for the sequence-to-sequence model. This local information is then concatenated together with the global graph context and fed into a deep stack of fully connected relu layers with residual connections between each layer.  The result is fed into a softmax, which selects the label from the (win, lose, or tie) outcomes.  

\subsubsection{P-Network}

The P-Network architecture is shown in Figure~\ref{fig:p-network}.  It computes the probability of assigning each of the valid color choices to the current vertex, using the global graph context, the problem context, and local context for each possible color. 
We draw inspiration from pointer-networks~\cite{pointer-networks} and represent colors with the embeddings of vertices that have previously been assigned the same colors.  In this analogy, the P-Network selects a pointer to a previously colored node rather than directly predicting a possible color.  However, our approach is different than pointer networks because it considers a set of pointers at a time with the same color rather than a single pointer.  It is also different because it considers a fixed set of possible colors instead of all previously encountered vertices.  These changes are important to exploit locality among nodes with the same color, boosting accuracy, and to bound the computational requirement for very large graphs to linear time in the order of the graph rather than quadratic time for pointer networks.  
In order to support a dynamic number of possible colors, each color is first processed independently, producing an unnormalized score.  This score is then post processed by a sequence-to-sequence model that incorporates dependencies on the other possible colors.  The final scores are normalized by a softmax operation.  

\subsection{High Performance Training System}

The computational requirements of training our networks on even a single large graph are vast.  In order to perform experiments quickly, we built a highly optimized training system.  The system uses tightly integrated compute nodes composed of commodity off-the-shelf GPUs, high performance networking, and high bandwidth IO system designed to serve training data.  


This system is organized into \textbf{AI POD}s.
Dense compute nodes with 10 1080Ti or 2080Ti GPUs are connected using FDR Infiniband.  A high performance IO node uses an SSD array to serve training data to the compute nodes at up to 7 GB/s.  Collectively the POD can sustain 3.12 single-precision PFLOP/s when running a single large training job.  When configured using 2080Ti GPUs, a single pod can sustain 32.1 mixed-precision PFLOP/s. In total we use up to 300 GPUs for training. 


We use data-parallelism for training on multiple GPUs using synchronous stochastic gradient descent with MPI for communication.  FastColorNet is designed to have high enough compute intensity to enable high GPU efficiency with a small batch size of 1-4 per GPU.  Using the entire AI POD requires a maximum batch size of 1200. The MCTS is completely data-parallel, with each data-parallel worker generating independent samples from different graphs.  Data parallelism is also used within a single GPU for inference to batch model evaluations for different graphs.  

\section{Empirical Analysis of Results}
\label{sec:results}

We applied this reinforcement learning process to train FastColorNet on a diverse set of commonly studied graph coloring problems, including synthetic Erdős–Rényi (ER)~\cite{erdos-renyi} and Watts–Strogatz (WS)~\cite{watts-strogatz} (sometimes called small-world) graphs, as well as the SuiteSparse collection~\cite{suite-sparse}.  Graph sizes varied from 32 vertices to 10 million vertices.

A single 1080Ti GPU was able to train FastColorNet on about 300,000 mini-batches of 4 moves in one day for a medium size graph with 512 vertices.   We use Adam optimizer for training with a fixed learning rate of 0.001, and did not use learning rate annealing or explicit regularization. We experimented with a variety of architectures for different datasets.  The best performing model architectures are described in the Appendix.

%







\begin{figure}
  \begin{minipage}{0.45\columnwidth}
    \includegraphics[width=\columnwidth]{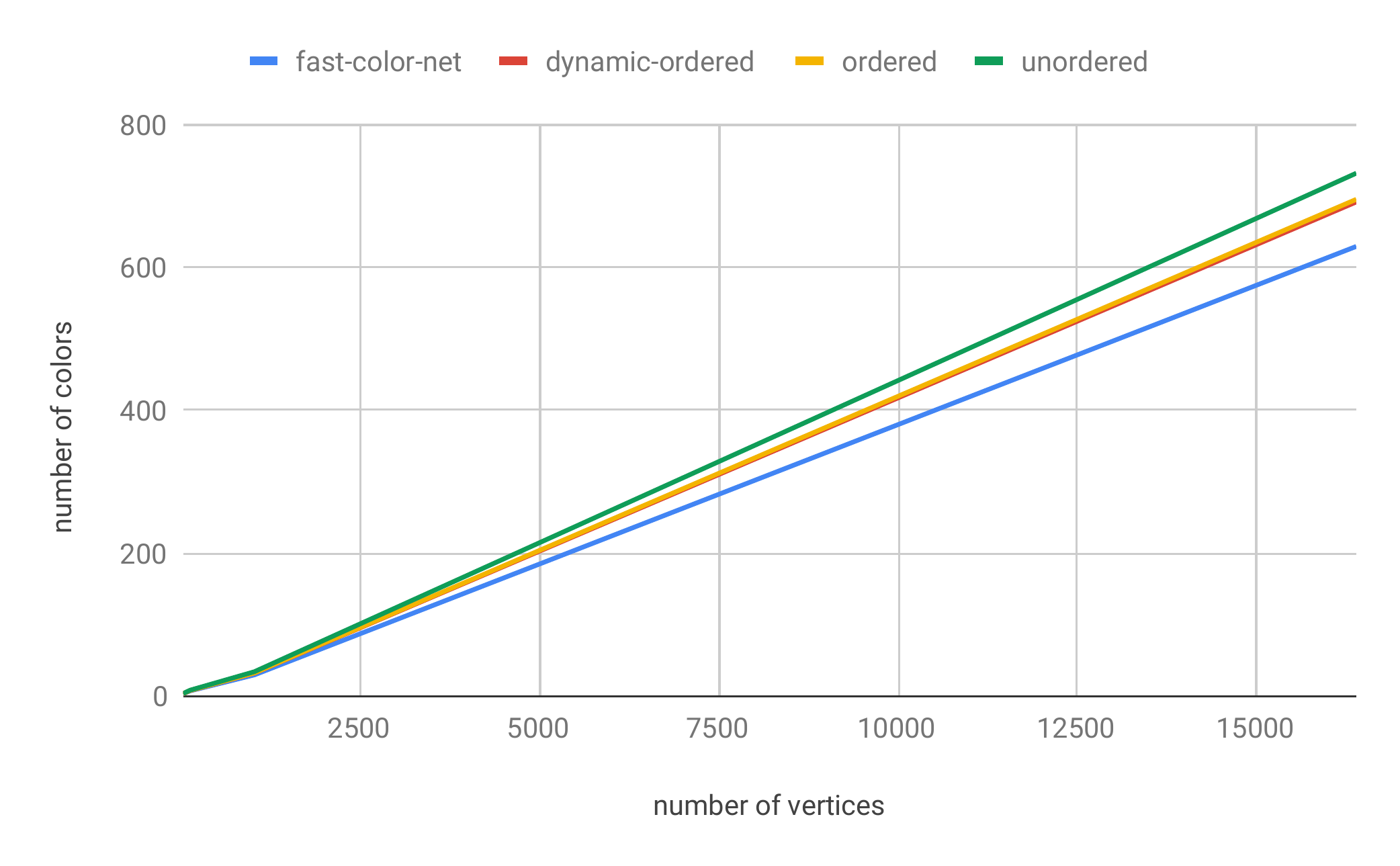}
    \caption{The number of colors used on the WS graph test sets as a function of vertex count.} 
    \label{fig:color-scaling}
  \end{minipage}
  \hfill
  \begin{minipage}{0.5\columnwidth}
    \includegraphics[width=\columnwidth]{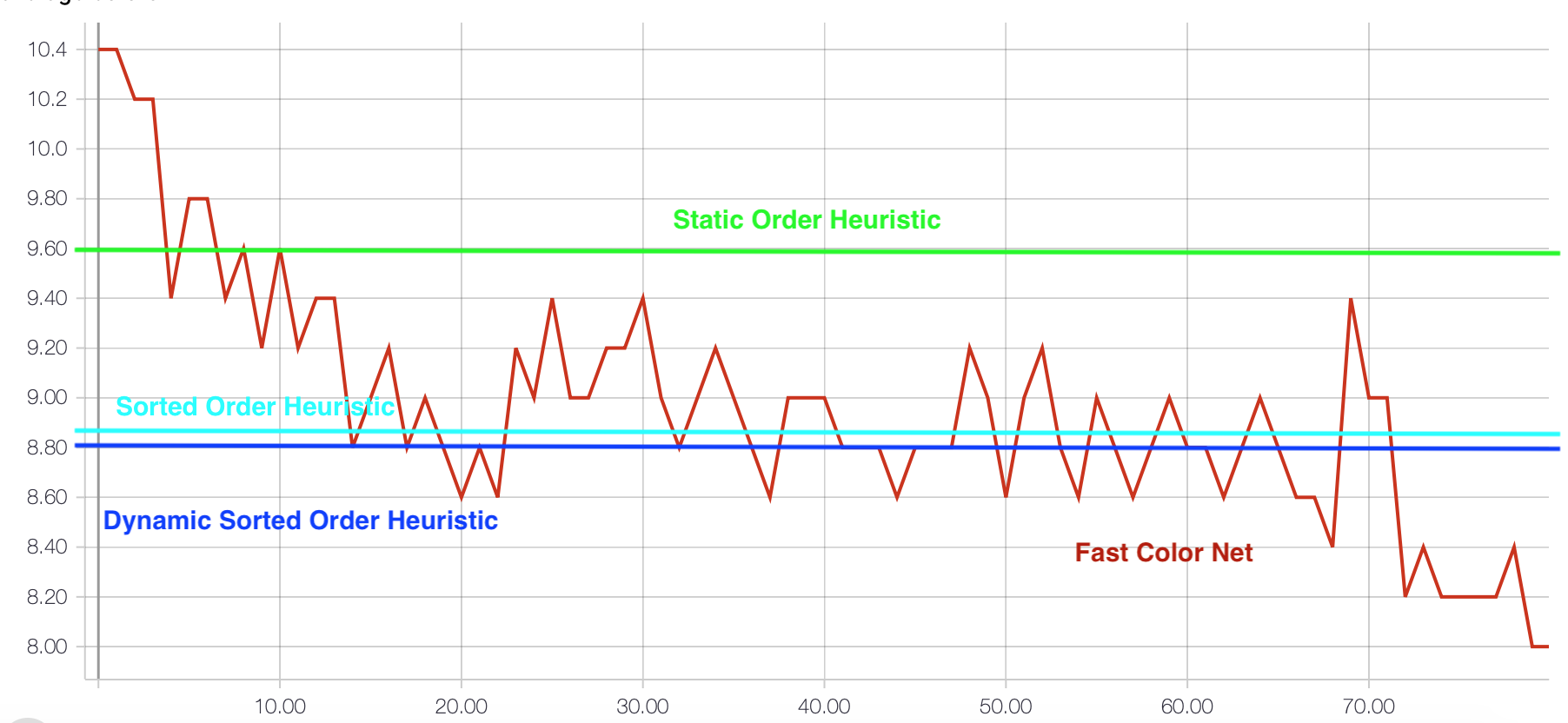}
    \caption{Improvement in learned heuristic with more training.}
    \label{fig:more-training}
  \end{minipage}
\end{figure}


To assess performance over a range of graph sizes, 
Figure~\ref{fig:color-scaling} shows how the number of required colors grows with the graph size for the small-world test set
.
FastColorNets in this experiment were trained on similar small-world graphs using the same graph generator parameters.  FastColorNet significantly improves on the best baseline.  



To explore how learning policies improve with more computation in training, Figure~\ref{fig:more-training} shows the performance improvement of FastColorNet on ER graphs as a function of the number of policy iteration steps.  Performance matches the best heuristic in under 20 steps, and significantly exceeds it in under 100 steps.  





\begin{table}[ht]
\begin{center}
  \caption{The average number of colors across our test sets. FCN is our FastColorNet architecture.
  SS means SuiteSparse (CIR: circuits, LP: linear programming, FE: finite-element).  FCN-train represents performance when a graph present in the training set is evaluated on, FCN-test uses a model trained on the same type of graph, and FCN-gen tests generalization performance of a model trained on random graphs of many sizes.}
\label{table:test-set-results}
  \resizebox{\textwidth}{!}{
\begin{tabular}{|l|l|l|l|l|l|l|l|l|l|l|}
\hline
                                                Dataset  & ER-1K          & WS-1K         & ER-16K          & WS-16K          & ER-10M          & WS-10M         & SS-CIR  & SS-LP         & SS-Web       & SS-FE         \\ \hline
\multicolumn{1}{|l|}{Unordered}                   & 34.3           & 59.2          & 732.8           & 265.35          & 42923        & 16415	          & 4.2          & 4.25          & 3.75         & 4.85          \\ 
\multicolumn{1}{|l|}{Ordered}                     & 32.45          & 57.35         & 715.2           & 261.8           & 40347        & 15922 & 3.15         & \textbf{2.95} & 2.6          & 4.05          \\ 
\multicolumn{1}{|l|}{Dynamic}             & 32.2           & 57.15         & 708.5           & 261.2           & 37524 & 15843 & 3.55         & 3.15          & 2.7          & 4.25          \\ \hline
\multicolumn{1}{|l|}{FCN-train} & \textbf{29.58} & \textbf{52.5} & \textbf{660.19} & \textbf{237.03} & \textbf{35362} & \textbf{14924} & \textbf{3.0} & \textbf{2.95} & \textbf{2.4} & \textbf{3.75} \\ 
\multicolumn{1}{|l|}{FCN-test}           & 31.7           & 56.59         & 702.57          & 258.3 & 37849 & 15023          & 3.1          & \textbf{2.95} & 2.55         & 4.1           \\ 
\multicolumn{1}{|l|}{FCN-gen}     & 33.9           & 57.66         & 708.13          & 267.53  & 43415 & 17262         & 4.15          & 4.3           & 3.7          & 4.95          \\ \hline
\end{tabular}
}
\end{center}
\redVspace
\redVspace
\end{table}

Table~\ref{table:test-set-results} shows the performance of FastColorNet on test sets of different sizes from different application domains.  Test sets have 20 graphs each, representing 32K-16M moves. 
The FCN-train results show that our reinforcement learning pipeline can find solutions that improve on those found by heuristics by $5\%-10\%$.  
FCN-test shows generalization performance to new graphs from the same domain (e.g. from circuits, linear programming, etc).   Performance is also $1\%-2\%$ better than the heuristics, showing the potential to improve performance through domain specialization.  FCN-gen is trained on random graphs. It improves on the unordered heuristic (it itself is unordered), but loses to the other heuristics.  
We expect that there is actually more headroom to further improve these results with more training.  Additionally, SuiteSparse domains actually exhibit a fair amount of diversity, e.g. the circuit category includes circuit layout problems as well as electrical simulation of circuits.





\section{Related Work}

\textbf{Deep RL for Games}.
Reinforcement learning has been researched for decades for games~\cite{temperal-difference-go, td-gammon}.
Mnih et al. proposed DQN, a combination of deep neural network and q-learning with experience replay, to achieve human-level performance on Atari games~\cite{dqn-atari}.
More recently, Deepmind published a series of AlphaGo algorithms for more complex game Go and defeated human experts~\cite{alpha-go, alpha-go-zero}.
They apply MCTS to explore the large MDP state space while balancing explore and exploit by using UCB for decision selection.
In addition, alpha-beta-pruning~\cite{alpha-beta-pruning} is also adopted to reduce tree search space.
Our approach for graph coloring takes the similar one of AlphaGoZero.
Furthermore, in order to learn for much bigger problems compared to the relatively small one of Go, we apply other innovations to make our solution scalable and computational efficiently.

\textbf{RL/ML for Combinatorial Optimization}.
Recently, reinforcement learning has been applied for combinatorial optimization.
Bello et al. combined pointer networks~\cite{pointer-networks} with actor and critic network to optimize Traveling Salesman Problem (TSP)~\cite{rl-tsp}, which does not make good use of graph structure and is not generalized to arbitrary size graphs. 
In~\cite{structure2vec-q-learning}, a Q-learning framework is introduced for greedy algorithms to learn over MVC, MAXCUT and TSP problems using structure2vec~\cite{message-passing-embeddings} graph embedding.
This algorithm cannot be directly applied to graph coloring since the reward design and state representation for colored graphs is non trivial in its problem formulation.
Both algorithms only evaluated on small graphs and are not scalable to big graphs, which are typical in real applications.
In contrast, our approach is easy to scale to train on bigger graphs to solve bigger problems.


\textbf{Graph Coloring}.
Graph coloring are being studied for several decades due to its usefulness in many practical applications, including linear algebra,
parallel computing, resource assignment and register allocation 
\cite{naumann2012combinatorial}. Graph coloring problem is known as NP-Hard and so its approximation is \cite{gebremedhin2005color}. 
Heuristics are therefore been used widely, which includes greedy
algorithms, finding independent sets, local search and population
based algorithms \cite{johnson2012color02}. An in-depth 
experimental study has been presented in \cite{gebremedhin2013colpack},
which demonstrated that greedy coloring algorithm with
appropriate vertex ordering gives close to optimal coloring
in reasonable time on wide range of applications. In this paper,
we used the implementation of \cite{gebremedhin2013colpack} 
when comparing our reinforcement based technique. 
Several variations of graph coloring (e.g. 
star and acyclic coloring) and heuristics algorithms are 
studied in 
\cite{gebremedhin2005color, gebremedhin2009efficient, gebremedhin2007new}.


\section{Conclusions}

Our results demonstrate that deep reinforcement learning can be applied to large scale problems of clear commercial importance: such as the well-known graph coloring problem.
The MCTS+UCB algorithm used in AlphaGoZero defeated state of the art heuristics by a large margin.
The learned heuristics generalized to new graphs in the same application domain, and are fast enough to color very large graphs.




\let\oldthebibliography\thebibliography
\let\endoldthebibliography\endthebibliography
\renewenvironment{thebibliography}[1]{
  \begin{oldthebibliography}{#1}
    \setlength{\itemsep}{1em}
    \setlength{\parskip}{0em}
}
{
  \end{oldthebibliography}
}
\bibliographystyle{unsrt}
{
  \footnotesize
  \bibliography{bibliography}

\begin{thebibliography}{10}

\bibitem{alpha-go-zero}
David Silver, Julian Schrittwieser, Karen Simonyan, Ioannis Antonoglou, Aja
  Huang, Arthur Guez, Thomas Hubert, Lucas Baker, Matthew Lai, Adrian Bolton,
  Yutian Chen, Timothy Lillicrap, Fan Hui, Laurent Sifre, George van~den
  Driessche, Thore Graepel, and Demis Hassabis.
\newblock Mastering the game of go without human knowledge.
\newblock {\em Nature}, 550:354--359, 2017.

\bibitem{structure2vec-q-learning}
Hanjun Dai, Elias~B. Khalil, Yuyu Zhang, Bistra Dilkina, and Le~Song.
\newblock Learning combinatorial optimization algorithms over graphs, 2017.

\bibitem{gcn-guided-tree-search}
Zhuwen Li, Qifeng Chen, and Vladlen Koltun.
\newblock Combinatorial optimization with graph convolutional networks and
  guided tree search.
\newblock In S.~Bengio, H.~Wallach, H.~Larochelle, K.~Grauman, N.~Cesa-Bianchi,
  and R.~Garnett, editors, {\em Advances in Neural Information Processing
  Systems 31}, pages 537--546. Curran Associates, Inc., 2018.

\bibitem{learned-auto-reason-heuristic}
Gil Lederman, Markus~N. Rabe, and Sanjit~A. Seshia.
\newblock Learning heuristics for automated reasoning through deep
  reinforcement learning.
\newblock {\em CoRR}, abs/1807.08058, 2018.

\bibitem{alpha-zero}
David Silver, Thomas Hubert, Julian Schrittwieser, Ioannis Antonoglou, Matthew
  Lai, Arthur Guez, Marc Lanctot, Laurent Sifre, Dharshan Kumaran, Thore
  Graepel, Timothy~P. Lillicrap, Karen Simonyan, and Demis Hassabis.
\newblock Mastering chess and shogi by self-play with a general reinforcement
  learning algorithm.
\newblock {\em CoRR}, abs/1712.01815, 2017.

\bibitem{smallest-hard-to-color-graph}
R.~Janczewski, M.~Kubale, K.~Manuszewski, and K.~Piwakowski.
\newblock The smallest hard-to-color graph for algorithm dsatur.
\newblock {\em Discrete Math.}, 236(1-3):151--165, June 2001.

\bibitem{naumann2012combinatorial}
Uwe Naumann and Olaf Schenk.
\newblock {\em Combinatorial scientific computing}.
\newblock CRC Press, 2012.

\bibitem{gebremedhin2005color}
Assefaw~Hadish Gebremedhin, Fredrik Manne, and Alex Pothen.
\newblock What color is your jacobian? graph coloring for computing
  derivatives.
\newblock {\em SIAM review}, 47(4):629--705, 2005.

\bibitem{zuckerman2006linear}
David Zuckerman.
\newblock Linear degree extractors and the inapproximability of max clique and
  chromatic number.
\newblock In {\em Proceedings of the thirty-eighth annual ACM symposium on
  Theory of computing}, pages 681--690. ACM, 2006.

\bibitem{ccatalyurek2012graph}
{\"U}mit~V {\c{C}}ataly{\"u}rek, John Feo, Assefaw~H Gebremedhin, Mahantesh
  Halappanavar, and Alex Pothen.
\newblock Graph coloring algorithms for multi-core and massively multithreaded
  architectures.
\newblock {\em Parallel Computing}, 38(10-11):576--594, 2012.

\bibitem{gebremedhin2013colpack}
Assefaw~H Gebremedhin, Duc Nguyen, Md~Mostofa~Ali Patwary, and Alex Pothen.
\newblock Colpack: Software for graph coloring and related problems in
  scientific computing.
\newblock {\em ACM Transactions on Mathematical Software (TOMS)}, 40(1):1,
  2013.

\bibitem{alpha-beta-pruning}
Donald~E Knuth and Ronald~W Moore.
\newblock An analysis of alpha-beta pruning.
\newblock {\em Artificial intelligence}, 6(4):293--326, 1975.

\bibitem{replay-buffer}
Long-Ji Lin.
\newblock Self-improving reactive agents based on reinforcement learning,
  planning and teaching.
\newblock {\em Machine learning}, 8(3-4):293--321, 1992.

\bibitem{embeddings-survey}
William~L Hamilton, Rex Ying, and Jure Leskovec.
\newblock Representation learning on graphs: Methods and applications.
\newblock {\em arXiv preprint arXiv:1709.05584}, 2017.

\bibitem{message-passing-embeddings}
Hanjun Dai, Bo~Dai, and Le~Song.
\newblock Discriminative embeddings of latent variable models for structured
  data.
\newblock In Maria~Florina Balcan and Kilian~Q. Weinberger, editors, {\em
  Proceedings of The 33rd International Conference on Machine Learning},
  volume~48 of {\em Proceedings of Machine Learning Research}, pages
  2702--2711, New York, New York, USA, 20--22 Jun 2016. PMLR.

\bibitem{sampled-embedding-training}
Hanjun Dai, Zornitsa Kozareva, Bo~Dai, Alex Smola, and Le~Song.
\newblock Learning steady-states of iterative algorithms over graphs.
\newblock In Jennifer Dy and Andreas Krause, editors, {\em Proceedings of the
  35th International Conference on Machine Learning}, volume~80 of {\em
  Proceedings of Machine Learning Research}, pages 1106--1114,
  Stockholmsmässan, Stockholm Sweden, 10--15 Jul 2018. PMLR.

\bibitem{truncated-back-prop}
Ronald~J Williams and Jing Peng.
\newblock An efficient gradient-based algorithm for on-line training of
  recurrent network trajectories.
\newblock {\em Neural computation}, 2(4):490--501, 1990.

\bibitem{pointer-networks}
Oriol Vinyals, Meire Fortunato, and Navdeep Jaitly.
\newblock Pointer networks.
\newblock In C.~Cortes, N.~D. Lawrence, D.~D. Lee, M.~Sugiyama, and R.~Garnett,
  editors, {\em Advances in Neural Information Processing Systems 28}, pages
  2692--2700. Curran Associates, Inc., 2015.

\bibitem{erdos-renyi}
Paul Erdos and Alfr{\'e}d R{\'e}nyi.
\newblock On the evolution of random graphs.
\newblock {\em Publ. Math. Inst. Hung. Acad. Sci}, 5(1):17--60, 1960.

\bibitem{watts-strogatz}
Duncan~J Watts and Steven~H Strogatz.
\newblock Collective dynamics of ‘small-world’networks.
\newblock {\em nature}, 393(6684):440, 1998.

\bibitem{suite-sparse}
Timothy~A Davis and Yifan Hu.
\newblock The university of florida sparse matrix collection.
\newblock {\em ACM Transactions on Mathematical Software (TOMS)}, 38(1):1,
  2011.

\bibitem{temperal-difference-go}
Nicol~N. Schraudolph, Peter Dayan, and Terrence~J Sejnowski.
\newblock Temporal difference learning of position evaluation in the game of
  go.
\newblock In J.~D. Cowan, G.~Tesauro, and J.~Alspector, editors, {\em Advances
  in Neural Information Processing Systems 6}, pages 817--824. Morgan-Kaufmann,
  1994.

\bibitem{td-gammon}
Gerald Tesauro.
\newblock Temporal difference learning and td-gammon.
\newblock {\em Commun. ACM}, 38(3):58--68, March 1995.

\bibitem{dqn-atari}
Volodymyr Mnih, Koray Kavukcuoglu, David Silver, Andrei~A. Rusu, Joel Veness,
  Marc~G. Bellemare, Alex Graves, Martin~A. Riedmiller, Andreas Fidjeland,
  Georg Ostrovski, Stig Petersen, Charles Beattie, Amir Sadik, Ioannis
  Antonoglou, Helen King, Dharshan Kumaran, Daan Wierstra, Shane Legg, and
  Demis Hassabis.
\newblock Human-level control through deep reinforcement learning.
\newblock {\em Nature}, 518(7540):529--533, 2015.

\bibitem{alpha-go}
David Silver, Aja Huang, Christopher~J. Maddison, Arthur Guez, Laurent Sifre,
  George van~den Driessche, Julian Schrittwieser, Ioannis Antonoglou, Veda
  Panneershelvam, Marc Lanctot, Sander Dieleman, Dominik Grewe, John Nham, Nal
  Kalchbrenner, Ilya Sutskever, Timothy Lillicrap, Madeleine Leach, Koray
  Kavukcuoglu, Thore Graepel, and Demis Hassabis.
\newblock Mastering the game of go with deep neural networks and tree search.
\newblock {\em Nature}, 529:484--503, 2016.

\bibitem{rl-tsp}
Irwan Bello, Hieu Pham, Quoc~V. Le, Mohammad Norouzi, and Samy Bengio.
\newblock Neural combinatorial optimization with reinforcement learning, 2016.

\bibitem{johnson2012color02}
David Johnson, Anuj Mehrotra, and Michael Trick.
\newblock Color02/03/04: Graph coloring and its generalizations, 2012.

\bibitem{gebremedhin2009efficient}
Assefaw~H Gebremedhin, Arijit Tarafdar, Alex Pothen, and Andrea Walther.
\newblock Efficient computation of sparse hessians using coloring and automatic
  differentiation.
\newblock {\em INFORMS Journal on Computing}, 21(2):209--223, 2009.

\bibitem{gebremedhin2007new}
Assefaw~H Gebremedhin, Arijit Tarafdar, Fredrik Manne, and Alex Pothen.
\newblock New acyclic and star coloring algorithms with application to
  computing hessians.
\newblock {\em SIAM Journal on Scientific Computing}, 29(3):1042--1072, 2007.

\end{thebibliography}
}

\pagebreak
\section{Appendix}
\label{sec:appendix}

\subsection{Dataset}

We use both synthetic random graphs and real-world graphs in our study.
We generate random graphs using both Erdos–Rényi (ER) and Watts-Strogatz (WS) models.
We implemented a random graph generator using ER model to generate graphs for training with edge probability of 0.5.
For WS graphs, also called small-world graphs, we set average degree $K$ to be 4 and rewiring probability $\beta$ to 0.5. 
For real-world graphs, we take them from SuiteSparse.

%
%
\subsection{Embedding Features}

\begin{table}[hbt]
  \vspace{-5mm}
  \begin{center}
    \caption{Graph embedding features and descriptions.}
    \label{table:graph-embedding}
    \resizebox{\textwidth}{!}{
      \begin{tabular}{l|c|c|l}
        \hline
        \textbf{Feature} & \textbf{Size} & \textbf{Encoding Format} & \textbf{Description} \\ \hline \hline
        degree & {32} & \multirow{4}{*}{\shortstack{One-Hot Encoding\\(Set position $\lfloor \frac{value}{max} \times size \rfloor$\\in $size$-long vector)}} & $value:$ node degree, $max:$ maximum degree of the graph\\ \cline{1-2} \cline{4-4}
        total-colors & {32} & & $value:$ total number of colors so far, $max:$ max number of colors\\ \cline{1-2} \cline{4-4}
        current-node & {32} & & $value:$ current node id, $max:$ total number of nodes \\ \cline{1-2} \cline{4-4}
        current-color & {32} & & $value:$ current color ID, $max:$ max number of colors \\ \hline
        possible-colors & {32} & Multi-Hot Encoding & $values:$ possible colors' IDs, $max:$ max number of colors \\ \hline
        node & {128} & learned & learn by the neural network described in Algorithm~\ref{alg:graph-embedding} \\ \hline
      \end{tabular}
    }
  \end{center}
  \vspace{-4mm}
\end{table}


Diverse types of embedding are used for learning the global and local graph structures, as well as the current state of the coloring progress.
We use multiple one-hot encoding and a multi-hot encoding for capturing the current node information and the current coloring situation.
These features are fed into the embedding network listed in Algorithm~\ref{alg:graph-embedding} to learn and produce the node embeddings. Details of the embedding setting are listed in Table~\ref{table:graph-embedding}.

\subsection{Hyperparameters}

Our configuration performed well across tasks.  The resnet stack in the V-Network in Figure~\ref{fig:v-network} included 3 layers with hidden dimensions of 512 units.  The resnet stack in the P-Network in Figure~\ref{fig:p-network} included 5 layers with hidden dimsions of 512 units.  Batch normalization was used between all residual layers.  The layers labeled seq2seq in are implemented with stacks of Conv1D with residual connections and batch normalization between all layers.  The number of Conv1D layers for all seq2seq operations includes 3 layers with 128 channels each and a filter size of 7.   We did not use regularization in any experiment.  We used the ADAM optimizer with a learning rate of between 0.0001 and 0.001.

%
%

\subsection{Baseline Graph Coloring Heuristics}

We implemented several well known graph coloring heuristics~\cite{gebremedhin2013colpack} for comparison.
When a vertex is selected for coloring, a valid color with smallest color ID is assigned to it.
The difference among these heuristics is the vertex order for coloring, described as follow:
\begin{wrapfigure}{r}{0.35\textwidth}
  \vspace{-3mm}
  \begin{center}
\includegraphics[width=0.35\textwidth]{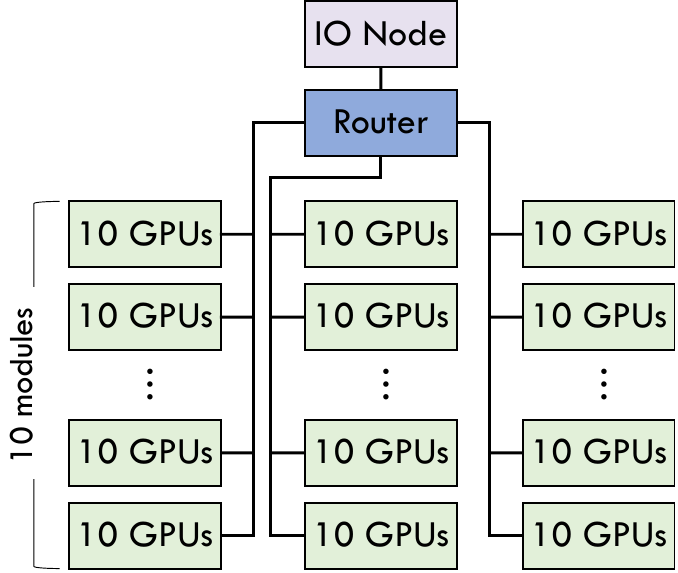}
\caption{The AI POD architecture.}
\label{fig:ai-accelerator-pod}
  \end{center}
  \vspace{-5mm}
\end{wrapfigure}
\begin{itemize}[noitemsep, topsep=0pt, leftmargin=2em]
  \item \textbf{Static-Ordered (Unordered):} This heuristic colors the vertices in the order of the node ID without any sorting.
  \item \textbf{Sorted-Ordered (Ordered):} We first sort the vertices from largest degree to smallest degree, then color the node with the sorted order.
  \item \textbf{Dynamic-Sorted-Ordered (Dynamic-Ordered):} In this heuristic, the order of vertices for coloring is based on the dynamic degree. Initially, a vertex's dynamic degree is same as its degree. At any time, a uncolored vertex with the largest dynamic degree is selected for coloring, then all is neighbors' dynamic degrees decrease by one.
\end{itemize}

\subsection{AI POD System}

Figure~\ref{fig:ai-accelerator-pod} shows the system configuration of the high-performance AI PODs we use in the training task.
It contains up to 300 tightly integrated GPUs with high bandwidth access to training data.


\end{document}